\documentclass{article}

\usepackage[square, numbers]{natbib}
\usepackage[preprint]{report_2019_JH}

\usepackage[utf8]{inputenc}         
\usepackage[T1]{fontenc}            
\usepackage{url}                    
\usepackage{booktabs}               
\usepackage{amsfonts}               
\usepackage{nicefrac}               
\usepackage{microtype}              

\usepackage{float}
\usepackage{adjustbox}
\usepackage{amsmath}
\usepackage{amssymb}
\usepackage{mathtools}
\usepackage{array}
\usepackage{enumitem}               
\usepackage{times}
\usepackage{longtable}
\usepackage{multirow}
\usepackage{xspace}
\usepackage{pgfgantt}
\usepackage{setspace}
\usepackage{subcaption}             
\usepackage{hyperref}               

\DeclareRobustCommand{\eg}{e.g.\@\xspace}
\DeclareRobustCommand{\ie}{i.e.\@\xspace}

\makeatletter
\DeclareRobustCommand{\etc}{%
    \@ifnextchar{.}%
    {etc}%
    {etc.\@\xspace}%
}

\DeclareRobustCommand\onedot{\futurelet\@let@token\@onedot}
\def\@onedot{\ifx\@let@token.\else.\null\fi\xspace}
\def\etal{\emph{et al}\onedot}
\makeatother

\newcolumntype{L}[1]{>{\raggedright\arraybackslash}p{#1}}
\newcolumntype{C}[1]{>{\centering\arraybackslash}p{#1}}

\newcolumntype{`}{>{\global\let\currentrowstyle\relax}}
\newcolumntype{~}{>{\currentrowstyle}}
\newcommand{\rowstyle}[1]{\gdef\currentrowstyle{#1}#1\ignorespaces }
\newcommand{\rbf}{\rowstyle{\bfseries} \boldmath}           
\newcommand{\tbf}[1]{\textbf{#1}}

\newcommand{\GradualPruning}{Gradual}
\newcommand{\BlindPruning}{Hard (class-blind)}
\newcommand{\UniformPruning}{Hard (class-uniform)}
\newcommand{\DistPruning}{Hard (class-distribution)}
\newcommand{\Proposed}{\emph{Our proposed}}

\newcommand{\NNZ}{p_{nnz}}
\newcommand{\TotalParams}{p_{total}}
\DeclareMathOperator{\sample}{z}
\DeclareMathOperator{\bern}{Bern}
\DeclareMathOperator{\round}{Round}
\DeclareMathOperator{\sigmoid}{\sigma}
\DeclareMathOperator{\RNN}{RNN}

\DeclareMathOperator{\Softmax}{Softmax}
\DeclareMathOperator{\SoftAttention}{SoftAtt}

\DeclarePairedDelimiter\abs{\lvert}{\rvert}%
\DeclarePairedDelimiter\norm{\lVert}{\rVert}%
\DeclarePairedDelimiter\set\{\}%

\title{Image Captioning with \\ Sparse Recurrent Neural Network}
\author{Jia Huei Tan\thanks{\{tanjiahuei@siswa.um.edu.my\}}\,\,\,\,\,\,Chee Seng Chan\thanks{\{cs.chan@um.edu.my\}} \\ Center of Image and Signal Processing,\\
        Faculty of Computer Science and Technology, \\ University of Malaya, 50603 Kuala Lumpur \\ Malaysia \\
        \And
        Joon Huang Chuah\thanks{\{jhchuah@um.edu.my\}} \\
        Department of Electrical Engineering, \\ Faculty of Engineering, \\ University of Malaya, \\ 50603 Kuala Lumpur, Malaysia \\
        }

\begin{document}

\maketitle

\begin{abstract}
Recurrent Neural Network (RNN) has been widely used to tackle a wide variety of language generation problems and are capable of attaining state-of-the-art (SOTA) performance. However despite its impressive results, the large number of parameters in the RNN model makes deployment in mobile and embedded devices infeasible. Driven by this problem, many works have proposed a number of pruning methods to reduce the sizes of the RNN model. In this work, we propose an end-to-end pruning method for image captioning models equipped with visual attention. Our proposed method is able to achieve sparsity levels up to $97.5\%$ without significant performance loss relative to the baseline ($\sim 2\%$ loss at $40 \times$ compression after fine-tuning). Our method is also simple to use and tune, facilitating faster development times for neural network practitioners. We perform extensive experiments on the popular MS-COCO dataset in order to empirically validate the efficacy of our proposed method.
\end{abstract}


\section{Introduction}
\label{sec: Introduction}

Automatically generating a caption that describes an image, a problem known as image captioning, is a challenging problem where computer vision (CV) meets natural language processing (NLP). 
A well performing model not only has to identify the objects in the image, but also capture the semantic relationship between them, general context and the activities that they are involved in. Lastly, the model has to map the visual representation into a fully-formed sentence in a natural language such as English.

A good image captioning model can have many useful applications, which include helping the visually impaired to better understand the web contents, providing descriptive annotations of website contents, and enabling better context-based image retrieval by tagging images with accurate natural language descriptions. 

Driven by user privacy concerns and the quest for lower user-perceived latency, deployment on edge devices away from remote servers is required. As edge devices usually have limited battery capacity and thermal limits, this presents a few key challenges in the form of storage size, power consumption and computational demands \cite{zhu2017prune}. 

For models incorporating RNNs, on-device inference is often memory bandwidth-bound. As RNN parameters are fixed at every time step, parameter reading forms the bulk of the work \cite{narang2017exploring,zhu2017prune}. As such, RNN pruning offers the opportunity to not only reduce the amount of memory access but also fitting the model in on-chip SRAM cache rather than off-chip DRAM memory, both of which dramatically reduce power consumption \cite{han2015deep,han2015learning}. Similarly, sparsity patterns for pruned RNNs are fixed across time steps. This offers the potential to factorise scheduling and load balancing operations outside of the loop and enable reuse \cite{narang2017exploring}. Lastly, pruning allows larger RNNs to be stored in memory and trained \cite{narang2017exploring,diamos2016persistent}

In this work, we propose a one-shot end-to-end pruning method to produce very sparse image captioning decoders (up to $97.5\%$ sparsity) while maintaining good performance relative to the dense baseline model as well as competing methods. We detail our contributions in the following section (Sec. \ref{subsec: Our contribution}).

\begin{figure}[t]
    \begin{center}
        \includegraphics[keepaspectratio=true, scale=0.35]{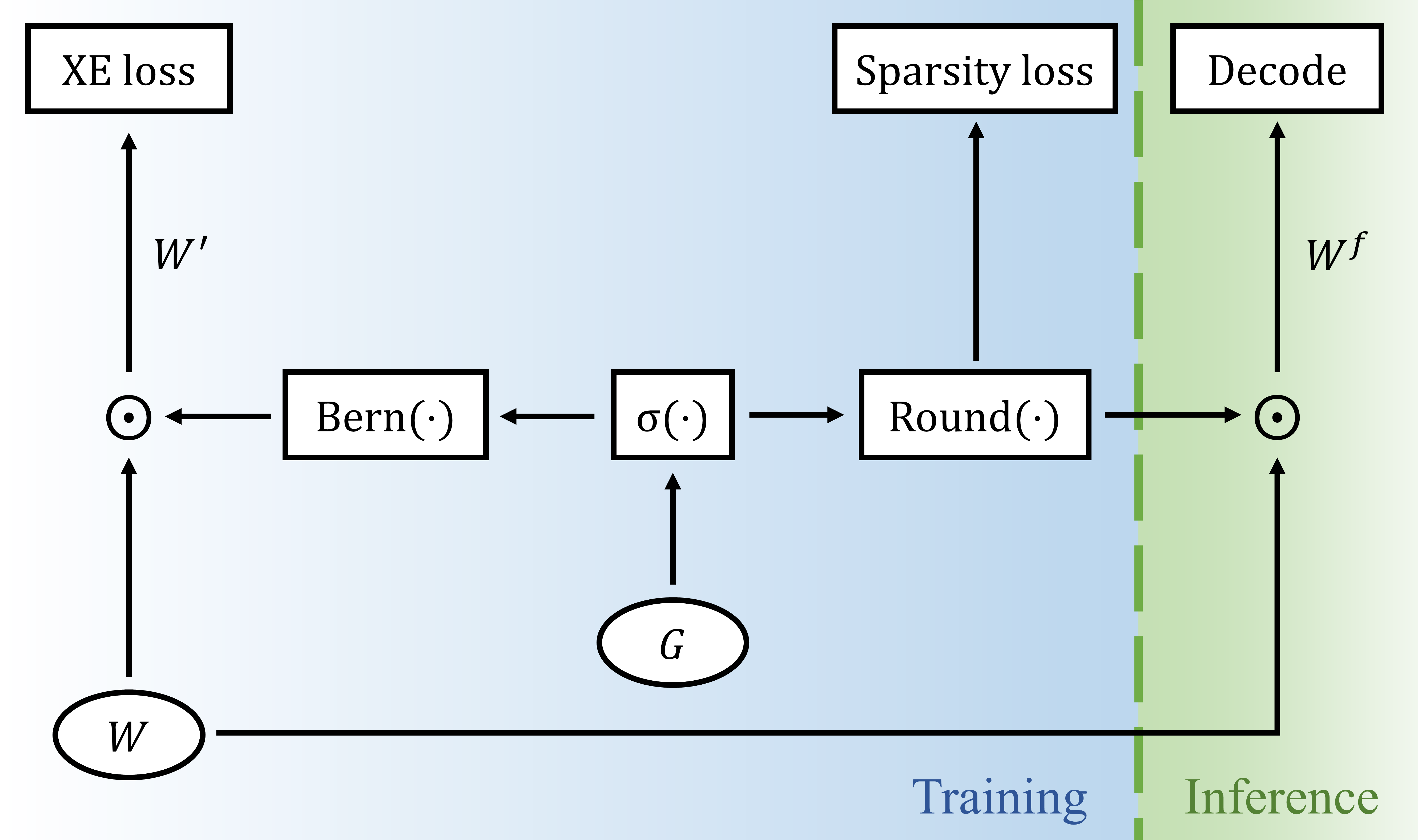}
    \end{center}
    \caption{An overview of our proposed end-to-end pruning method. ``XE'' denotes cross-entropy. More details can be found in Sec. \ref{subsec: End-to-end pruning}.}
    \label{fig: Overview}
\end{figure}


\section{Related Works}
\label{sec: Related Works}

Our work is most related to the current research on model pruning, particularly involving generative language RNNs. This section reviews the most relevant works on this topic.

\subsection{Model pruning}
\label{subsec: Model pruning}

Modern neural networks that provide good performance tend to be large and overparameterised, fuelled by observations that larger \cite{bengio2006convex,hinton2015distilling,zhang2016understanding} networks tend to be easier to train. This in turn drives numerous efforts to reduce model size using techniques such as weight pruning and quantisation \cite{courbariaux2015binaryconnect,hubara2017quantized,rastegari2016xnor}.

Early works like \cite{lecun1990optimal} and \cite{hassibi1993optimal} explored pruning by computing the Hessian of the loss with respect to the parameters in order to assess the saliency of each parameter. Other works involving saliency computation include \cite{mozer1989skeletonization} and \cite{karnin1990simple} where sensitivity of the loss with respect to neurons and weights are used respectively. On the other hand, works such as \cite{chauvin1989back,ishikawa1996structural} directly induce network sparsity by incorporating sparsity-enforcing penalty terms into the loss function.

Most of the recent works in network pruning focused on vision-centric classification tasks using Convolutional Neural Networks (CNNs) and occasionally RNNs. Techniques proposed include magnitude-based pruning \cite{han2015deep,han2015learning,guo2016dynamic} and variational pruning \cite{kingma2015variational,molchanov2017variational,dai2018compressing}. Among these, magnitude-based weight pruning have become popular due to their effectiveness and simplicity. 
Most notably, \cite{han2015deep} employed a combination of pruning, quantization and Huffman encoding resulting in massive reductions in model size without affecting accuracy.
While unstructured sparse connectivity provides reduction in storage size, it requires sparse General Matrix-Matrix Multiply (GEMM) libraries such as cuSPARSE and SPBLAS in order to achieve accelerated inference.
Motivated by existing hardware architectures optimised for dense linear algebra, many works propose techniques to prune and induce sparsity in a structured way in which entire filters are removed \cite{li2016pruning,luo2017thinet,yu2018nisp}.

On the other hand, works extending connection pruning to RNN networks are considerably fewer \cite{see2016compression,narang2017exploring,zhu2017prune,lee2018snip}. See \etal{} \cite{see2016compression} first explored magnitude-based pruning applied to deep multi-layer neural machine translation (NMT) model with Long-Short Term Memory (LSTM) \cite{hochreiter1997long}. In their work, three pruning schemes are evaluated which include class-blind, class-uniform and class-distribution. Class-blind pruning was found to produce the best result compared to the other two schemes. Narang \etal{} \cite{narang2017exploring} introduced a gradual magnitude-based pruning scheme for speech recognition RNNs whereby all the weights in a layer less than some chosen threshold are pruned. Gradual pruning is performed in parallel with network training while pruning rate is controlled by a slope function with two distinct phases. This is extended by Zhu and Gupta \cite{zhu2017prune} who simplified the gradual pruning scheme with reduced hyperparameters.

\subsection{Our contribution}
\label{subsec: Our contribution}

Our proposed end-to-end pruning method possesses three main qualities:

\begin{enumerate}[label=\roman*)]
    \item {
    \tbf{Simple and fast.} Our approach enables easy pruning of the RNN decoder equipped with visual attention, whereby the best number of weights to prune in each layer is automatically determined. Compared to works such as \cite{zhu2017prune,narang2017exploring}, our approach is simpler with $1$ to $2$ hyperparameters versus $3$ to $4$ hyperparameters. Our method also does not rely on reinforcement learning techniques such as in the work of \cite{he2018amc}. Moreover, our method applies pruning to all the weights in the RNN decoder and does not require special considerations to exclude pruning from certain weight classes. Lastly our method completes pruning in a single-shot process rather than requiring iterative train-and-prune process as in \cite{frankle2018lottery,dai2018grow,dai2019nest,yu2019playing}.
    }
    \item {
    \tbf{Good performance-to-sparsity ratio enabling very high sparsity.} Our approach achieves good performance across sparsity levels from $80\%$ up until $97.5\%$ ($40\times$ reduction in Number of Non-zeros (NNZ) parameters). This is in contrast with competing methods \cite{zhu2017prune,see2016compression} where there is a significant performance drop-off starting at sparsity level of $90\%$.
    }
    \item {
    \tbf{Easily tunable sparsity level.} Our approach provides a way for neural network practitioners to easily control the level of sparsity and compression desired. This allows for model solutions that are tailored for each particular scenario. In contrast, while the closely related works of \cite{srinivas2017training,louizos2017learning} also provide good performance with the incorporation of gating variables, there is not a straightforward way of controlling the final sparsity level. In their works, regularisers such as bi-modal, $l_2$ , $l_1$ and $l_0$ regulariser are used to encourage network sparsity. Their work also only focuses on image classification using CNNs.
    }
\end{enumerate}

While there are other works on compressing RNNs, most of the methods proposed either comes with structural constraints or are complementary to model pruning in principle. Examples include using low-rank matrix factorisations \cite{lu2016learning,kusupati2018fastgrnn}, product quantisation on embeddings \cite{shi2018structured}, factorising word predictions into multiple time steps \cite{li2016lightrnn,parameswaran2017exploring,tan2019comic}, and grouping RNNs \cite{gao2018efficient}.

Lastly, another closely related work by \cite{dai2018grow} also incorporated model pruning into image captioning. However we note three notable differences: 1) their work is focused on proposing a new LSTM cell structure named the \emph{H-LSTM}; 2) their work utilises the grow-and-prune (GP) method \cite{dai2019nest} which necessitates compute and time expensive iterative pruning; and 3) the compression figures stated are calculated based on the size of the LSTM cells instead of the entire decoder.


\section{Proposed Method}
\label{sec: Proposed Method}

Our proposed method involves incorporating learnable gating parameters into regular image captioning framework. We denote weight, bias and gating matrices as $W$, $B$ and $G$ respectively. For a model with $L$ layers, the captioning and gating parameters are denoted as $\theta$ and $\phi$ such that $\theta = \set{W_{1:L}, B_{1:L}}$ and $\phi = \set{G_{1:L}}$.

As there are substantial existing works focusing on pruning CNNs, we focus our efforts on pruning generative RNNs. As such, we only prune the RNN decoder. All model size calculations in this work include only the decoder (including attention module) while the encoder (\ie{} CNN) is excluded.

\begin{figure*}[t]
    \begin{center}
        \begin{subfigure}{.45\linewidth}
            \centering
            \includegraphics[keepaspectratio=true, scale=0.2]{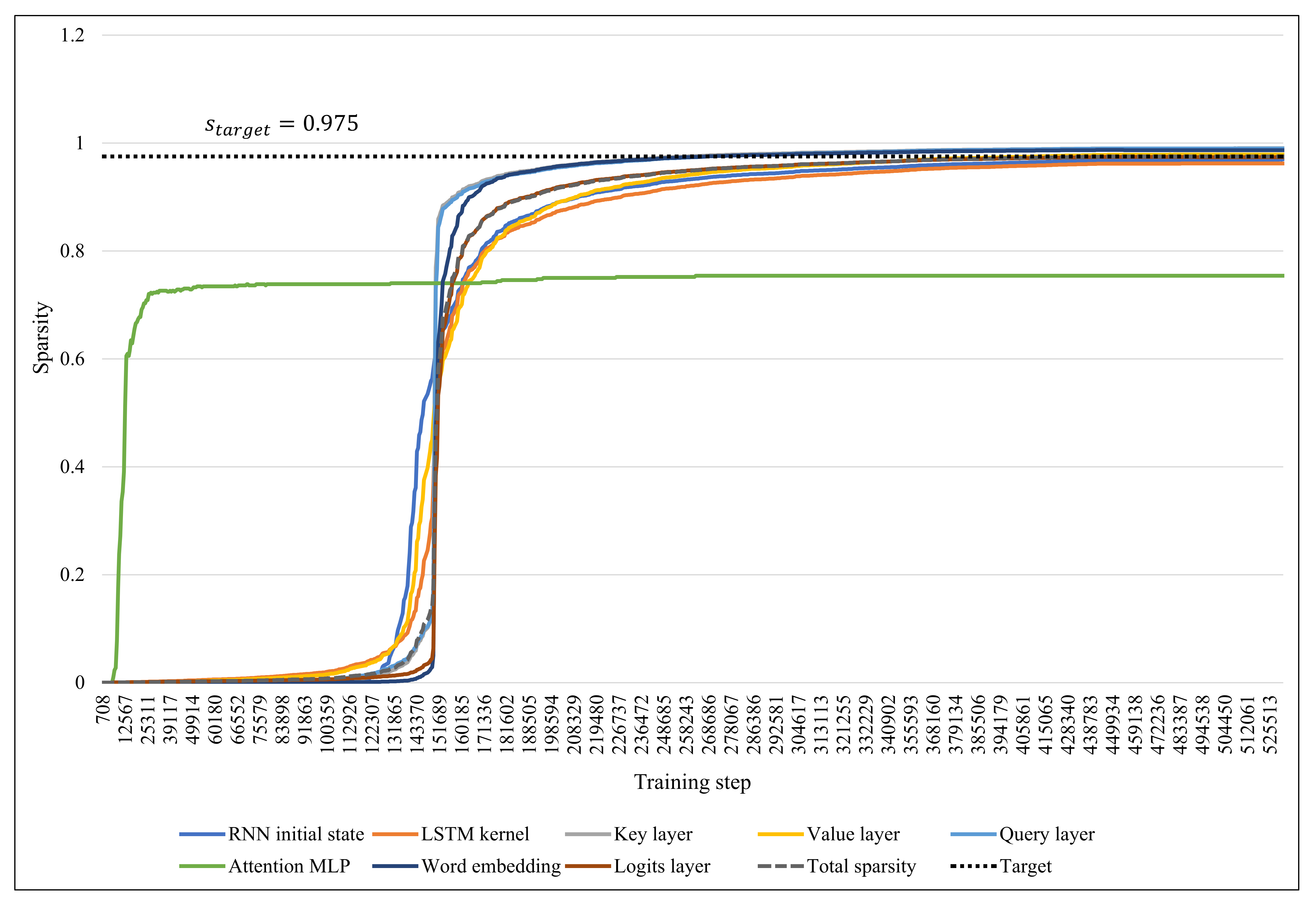}
            \caption{Sparsity level for different layers of the RNN decoder.}
            \label{fig: Sparsity progression}
        \end{subfigure}
        \hfill
        \begin{subfigure}{.45\linewidth}
            \centering
            \includegraphics[keepaspectratio=true, scale=0.05]{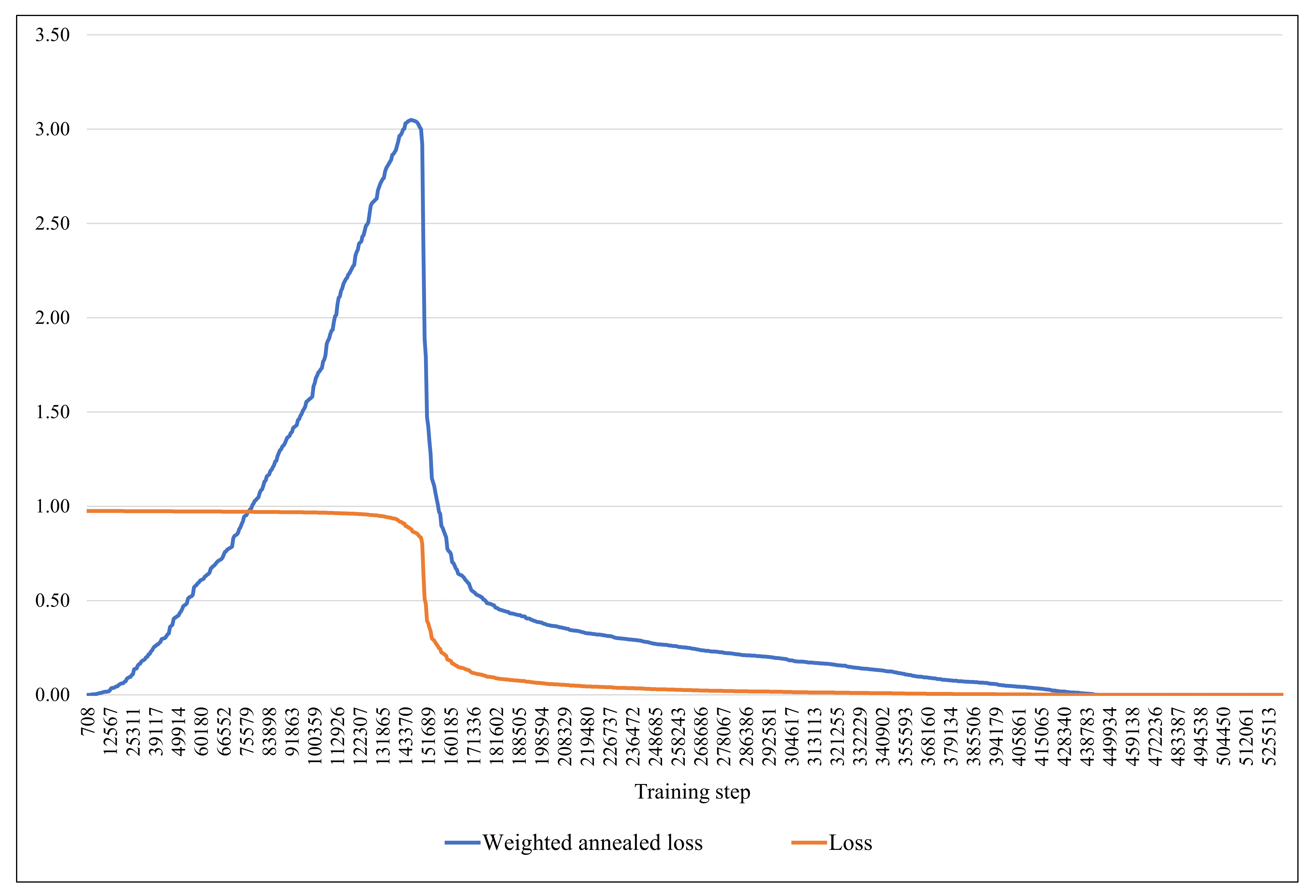}
            \caption{Sparsity loss $L_{s}$.}
            \label{fig: Sparsity loss}
        \end{subfigure}
    \end{center}
    \caption{Training progression of our proposed end-to-end pruning method. Best viewed in colour. Detailed explanation for (a) is given in Sec. \ref{subsec: Layer-wise sparsity}. In (b), ``Weighted annealed loss'' refers to $\lambda_{s} L_{s}$ in Eq. \ref{eq: final loss} while ``Loss'' refers to $L_{s}$ before applying cosine annealing in Eq. \ref{eq: sparsity loss}.}
    \label{fig: Training progression}
\end{figure*}

\subsection{Image captioning with visual attention}
\label{subsec: Image captioning}

Our image captioning framework of interest is a simplified variant of the \emph{Show, Attend and Tell} \cite{xu2015show} model which uses a single layer RNN network equipped with visual attention on the CNN feature map. It is a popular framework that forms the basis for subsequent state-of-the-art (SOTA) works on image captioning \cite{fu2016aligning,anderson2018bottom}. In this work, we employ LSTM and Gated Recurrent Unit (GRU) \cite{cho2014learning} as the RNN cell.

Suppose $\left\{S_0,\: ...\:, \:S_{T-1}\right\}$ is a sequence of words in a sentence of length $T$, the model directly maximises the probability of the correct description given an image $I$ using the following formulation:
\begin{equation}\label{eq: log-likelihood}
    \log{p}\left( S\,|\,I \right)=\sum_{t\,=\,0}^T\log{p}\left( S_t\,|\,I,\; S_{0\,:\,t-1}, \;c_t \right)
\end{equation}
\noindent where $t$ is the time step, $p\left( S_t\,|\,I,\; S_{0\::\:t-1},\; c_t \right)$ is the probability of generating a word given an image $I$, previous words $S_{0\::\:t-1}$, and context vector $c_t$. 

For a RNN network with $r$ units, the hidden state of RNN is initialised with the image embedding vector as follows:
\begin{equation}\label{eq: LSTM init}
    h_{t=-1} = W_I\,I_{embed} \:,\: m_{t=-1} = 0
\end{equation}
\noindent where $W_I \in\mathbb{R}\:^{r \times h}$ is a weight matrix and $h$ is the size of $I_{embed}$.

The attention function used in this work is \emph{soft-attention} introduced by \cite{bahdanau2014neural_iclr} and used in \cite{xu2015show}, where a multilayer perceptron (MLP) with a single hidden layer is employed to calculate the attention weights on a particular feature map. The context vector $c_t$ is then concatenated with previous predicted word embedding to serve as input to the RNN. Finally, a probability distribution over the vocabulary is produced from the hidden state $h_t$:

\begin{equation}\label{eq: RNN output}
    p_t = \Softmax \left( E_o\,h_t \right)
\end{equation}
\begin{equation}\label{eq: LSTM}
    h_t\:,\: m_t = \RNN\left(x_t \:,\: h_{t-1} \:,\: m_{t-1}\right)
\end{equation}
\begin{equation}\label{eq: RNN input}
    x_t = \left[ E_w\:S_{t-1},\; c_t \right]
\end{equation}
\begin{equation}\label{eq: Context}
    c_t = \SoftAttention \left( f \right)
\end{equation}
\noindent where $E_w \in\mathbb{R}\,^{q \times v}$ and $E_o \in\mathbb{R}\,^{v \times r}$ are input and output embedding matrices respectively; $p_t$ is the probability distribution over the vocabulary $V$; $m_t$ is the memory state; $x_t$ is the current input; $S_{t-1} \in\mathbb{R}\,^q$ is the one-hot vector of previous word; $c_t \in\mathbb{R}\,^a$ is the context vector; $f$ is the CNN feature map; and $\left[ \,, \right]$ is the concatenation operator. For GRU, all $m_{t}$ terms are ignored.

Finally, the standard cross-entropy loss function for the captioning model $\theta$ is given by:

\begin{equation}\label{eq: caption loss}
	L_{c} = - \sum_{t}^{T} \: \log p_{\,t} \left( S_{\,t} \right) + \lambda_{d} \, \norm{\theta}^2_2
\end{equation}

\subsection{End-to-end pruning}
\label{subsec: End-to-end pruning}

\noindent \tbf{Formulation.} Similar to \cite{zhu2017prune}, TensorFlow framework is extended to prune network connections during training. Inspired by the concept of learnable \emph{Supermasks} introduced by \cite{srinivas2017training,zhou2019deconstructing}, our proposed method achieves model pruning via learnable gating variables that are trained in an end-to-end fashion. An overview of our method is illustrated in Fig. \ref{fig: Overview}.

For every weight variable matrix $W$ to be pruned, we create a gating variable matrix $G$ with the same shape as $W$. This gating matrix $G$ functions as a masking mechanism that determines which of the parameter $w$ in the weight matrix $W$ participates in both forward-execution and back-propagation of the graph. 

To achieve this masking effect, we calculate the effective weight tensor as follows:

\begin{equation}\label{eq: weight masking}
	W^{'}_{l} = W_{l} \odot G^{b}_{l}
\end{equation}
\begin{equation}\label{eq: gating sample}
	G^{b}_{l} = \sample( \sigmoid( G_{l} ) )
\end{equation}

\noindent where $W_{l} , G_{l} \in\mathbb{R}^{D}$ are the original weight and gating matrices from layer $l$ with shape $D$; and superscript $\left(\cdot\right)^{b}$ indicates binary sampled variables. $\odot$ is element-wise multiplication; $\sigmoid(\cdot)$ is a point-wise function that transforms continuous values into the interval $(0, 1)$; and $\sample(\cdot)$ is a point-wise function that samples from a Bernoulli distribution. The composite function $\sample( \sigmoid( \cdot ) )$ thus effectively transforms continuous values into binary values.

Binary gating matrices $G^{b}$ can be obtained by treating $\sigmoid( G )$ as Bernoulli random variables. While there are many possible choices for the $\sigmoid$ function, we decided to use the logistic sigmoid function following \cite{bengio2013estimating} and \cite{zhou2019deconstructing}. To sample from the Bernoulli distribution, we can either perform a \emph{unbiased draw} or a \emph{maximum-likelihood} (ML) draw \cite{srinivas2017training}. Unbiased draw is the usual sampling process where a gating value $g \in (0, 1)$ is binarised to $1.0$ with probability $g$ and $0.0$ otherwise, whereas ML draw involves thresholding the value $g$ at $0.5$. In this work, we denote unbiased and ML draw using the sampling functions $\sample(\cdot) = \bern(\cdot)$ and $\sample(\cdot) = \round(\cdot)$ respectively. We back-propagate through both sampling functions using the \emph{straight-through estimator} \cite{bengio2013estimating} (\ie{} $\delta \sample(g) / \delta g = 1$).

Prior to training, all the gating variables are initialised to the same constant value $m$ while the weights and biases of the network are initialised using standard initialisation schemes (\eg{} Xavier \cite{glorot2010understanding}). During training, both sampling functions $\bern(\cdot)$ and $\round(\cdot)$ are used in different ways. To obtain the effective weight tensor used to generate network activations, we utilised $\bern(\cdot)$ to inject some stochasticity that helps with training and to mitigate the bias arising from the constant value initialisation. Thus the effective weight calculation becomes:

\begin{equation}\label{eq: weight masking bern}
	W^{'}_{l} = W_{l} \odot \bern( \sigmoid( G_{l} ) )
\end{equation}

To drive the sparsity level of gating variables $\phi$ to the user-specified level $s_{target}$, we introduce a regularisation term $L_{s}$. Consistent with the observations in the works of \cite{zhu2017prune} and \cite{yu2019playing}, we found that annealing the loss over the course of training produces the best result. Annealing is done using a cosine curve $\alpha$ defined in Eq. \ref{eq: sparsity anneal}. To ensure determinism when calculating sparsity, we use $\round(\cdot)$ to sample from $\sigmoid( G )$:

\begin{equation}\label{eq: sparsity loss}
    L_{s} = (1 - \alpha) \times \abs*{ \: s_{target} - \left( 1 - \frac{\NNZ}{\TotalParams} \right) \: }
\end{equation}
\begin{equation}\label{eq: sparsity anneal}
    \alpha = \frac{1}{2} \left( 1 + \cos \left( \frac{n \pi}{n_{max}} \right) \right)
\end{equation}
\begin{equation}\label{eq: nnz}
    \NNZ = \sum^{L}_{l\,=\,0}\sum^{J}_{j\,=\,0} \round( \sigmoid( g_{j,l} ) ) \:
\end{equation}

\noindent where $\NNZ$ is the number of NNZ gating parameters; $\TotalParams$ is the total number of gating parameters; $n$ and $n_{max}$ is the current and final training step respectively; $g_{j,l}$ is the gating parameter at position $j$ in the matrix $G_{l}$ from layer $l$; $L$ is the number of layers; and $J$ is the number of parameters in matrix $G_{l}$. The progression of sparsity loss $L_{s}$ as well as the sparsity levels of various layers in the decoder are illustrated in Fig. \ref{fig: Sparsity loss} and \ref{fig: Sparsity progression} respectively.

The final objective function used to train the captioning model $\theta$ with gating variables $\phi$ is:

\begin{equation}\label{eq: final loss}
    L \left(\, I,\, S,\, s_{target} \,\right) = L_{c} + \lambda_{s} L_{s}
\end{equation}

Intuitively, the captioning loss term $L_{c}$ provides supervision for learning of the saliency of each parameter where important parameters are retained with higher probability while unimportant ones are dropped more frequently. On the other hand, the sparsity regularisation term $L_{s}$ pushes down the average value of the Bernoulli gating parameters so that most of them have a value less than $0.5$ after sigmoid activation. The hyperparameter $\lambda_{s}$ determines the weightage of $L_{s}$. If $\lambda_{s}$ is too low, the target sparsity level might not be attained; whereas high values might slightly affect performance (see Sec. \ref{subsec: Ablation study}).

\noindent \tbf{Training and Inference.} The training process of the captioning model is divided into two distinct stages: decoder training and end-to-end fine-tuning. During the decoder training stage, we freeze the CNN parameters and only learn decoder and gating parameters by optimising the loss given in Eq. \ref{eq: final loss}. For the fine-tuning stage, we restore all the parameters $\theta$ and $\phi$ from the last checkpoint at the end of decoder training and optimise the entire model including the CNN. During this stage, $\bern(\cdot)$ is still used but all $\phi$ parameters are frozen.

After training is completed, all the weight matrices $W_{1:L}$ is transformed into sparse matrices by sampling from $G_{1:L}$ using $\round(\cdot)$, after which $G$ can be discarded. In other words, the final weights $W^{f}$ are calculated as:

\begin{equation}\label{eq: weight masking final}
	W^{f}_{l} = W_{l} \odot \round( \sigmoid( G_{l} ) )
\end{equation}


\begin{table*}[t]
    \caption{The effects of varying gating variable initialisation value $m$ on MS-COCO. Sparsity level is set to $0.8$. \tbf{Bold text} indicates best overall performance.}
    \label{table: Ablation: gating init}
    \begin{center}
    \begin{adjustbox}{max width=0.8\linewidth}
    \begin{tabular}{ `l  c ~c ~c ~c ~c ~c ~c ~c ~c }
        \toprule
        Gating init. value          && \multicolumn{8}{c}{MS-COCO test set scores} \\
                                       \cmidrule{3-10}
        \null                       && B-1  & B-2  & B-3  & B-4  & M    & R    & C    & S    \\
        
        \midrule
   \rbf $m = 5.0$                   && 71.6 & 54.8 & 41.4 & 31.4 & 24.6 & 52.8 & 94.4 & 17.5 \\
        $m = 2.5$                   && 71.3 & 54.5 & 41.1 & 31.1 & 24.4 & 52.5 & 93.1 & 17.4 \\
        $m = 0$                     && 71.3 & 54.3 & 40.8 & 30.6 & 24.4 & 52.6 & 92.4 & 17.3 \\
        $m = -2.5$                  && 70.8 & 53.8 & 40.2 & 30.1 & 24.1 & 52.1 & 91.1 & 17.0 \\
        $m = -5.0$                  && 70.5 & 53.5 & 39.8 & 29.5 & 23.6 & 51.8 & 88.0 & 16.5 \\
        
        \bottomrule
    \end{tabular}
    \end{adjustbox}
    \end{center}
\end{table*}

\begin{table*}[t]
    \caption{The effects of varying sparsity loss weightage $\lambda_{s}$ on MS-COCO. Sparsity target is set to $s_{target} = 0.9$. \tbf{Bold text} indicates best overall performance.}
    \label{table: Ablation: sparsity weight}
    \begin{center}
    \begin{adjustbox}{max width=0.8\linewidth}
    \begin{tabular}{ `l  c ~c  c ~c ~c ~c ~c ~c ~c ~c ~c }
        \toprule
        Gating init. value          && Sparsity && \multicolumn{8}{c}{MS-COCO test set scores} \\
                                                       \cmidrule{5-12}
        \null                       && \null    && B-1  & B-2  & B-3  & B-4  & M    & R    & C    & S    \\
        
        \midrule
        $\lambda_{s} = 1.0$         && 0.662    && 71.4 & 54.5 & 41.1 & 31.0 & 24.7 & 52.7 & 94.1 & 17.4 \\
        $\lambda_{s} = 2.0$         && 0.830    && 71.6 & 54.7 & 41.1 & 31.0 & 24.6 & 52.7 & 93.9 & 17.5 \\
   \rbf $\lambda_{s} = 5.0$         && 0.900    && 71.4 & 54.3 & 40.8 & 30.8 & 24.4 & 52.4 & 93.1 & 17.3 \\
        $\lambda_{s} = 10.0$        && 0.900    && 71.1 & 54.3 & 40.8 & 30.6 & 24.4 & 52.5 & 92.7 & 17.3 \\
        
        \bottomrule
    \end{tabular}
    \end{adjustbox}
    \end{center}
\end{table*}

\section{Experiment Setup}
\label{sec: Experiment Setup}

Unless stated otherwise, all experiments have the following configurations. We did not perform extensive hyperparameter search due to limited resources.

\subsection{Hyperparameters}
\label{subsec: Hyperparameters}

Models are implemented using TensorFlow r1.9. The image encoder used in this work is \emph{GoogLeNet (InceptionV1)} with batch normalisation \cite{szegedy2015going,ioffe2015batch} pre-trained on ImageNet \cite{deng2009imagenet}. The input images are resized to $256 \times 256$, then randomly flipped and cropped to $224 \times 224$ before being fed to the CNN. The attention function $\SoftAttention(\cdot)$ operates on the \emph{Mixed-4f} map $f \in\mathbb{R}\:^{196 \times 832}$. The size of context vector $c_t$ and attention MLP is set to $a = 512$. A single layer LSTM or GRU network with hidden state size of $r = 512$ is used. The word size is set to $q = 256$ dimensions.

The optimiser used for decoder training is Adam \cite{kingma2014adam_iclr}, with batch size of $32$. The initial learning rate (LR) is set to $1 \times 10^{-2}$, and annealed using the cosine curve $\alpha$ defined in Eq. \ref{eq: sparsity anneal}, ending at $1 \times 10^{-5}$. All models are trained for $30$ epochs. Weight decay rate is set to $\lambda_{d} = 1 \times 10^{-5}$. For fine-tuning, a smaller initial LR of $1 \times 10^{-3}$ is used and the entire model is trained for $10$ epochs. Captioning model parameters are initialised randomly using Xavier uniform initialisation \cite{glorot2010understanding}.

The input and output dropout rates for dense RNN are both set to $0.35$, while the attention map dropout rate is set to $0.1$. Following \cite{han2015learning,narang2017exploring}, a lower dropout rate is used for sparse networks where RNN and attention dropout rates are set to $0.11$ and $0.03$ respectively. This is done to account for the reduced capacity of the sparse models.

For fair comparison, we apply pruning to all weights of the captioning model for all of the pruning schemes. For our proposed method, we train the gating variables $\phi$ with a higher constant LR of $100$ without annealing, which is consistent with \cite{zhou2019deconstructing}. We found that LR lower than $100$ causes $\phi$ to train too slowly. We set $\lambda_{s}$ according to this heuristic: $\lambda_{s} = \max(5, 0.5 / (1-s_{target}))$. All gating parameters $\phi$ are initialised to a constant $m = 5.0$, see Sec. \ref{subsec: Ablation study} for other values.

For \emph{gradual pruning} \cite{zhu2017prune}, pruning is started after first epoch is completed and ended at the end of epoch $15$, following the general heuristics outlined in \cite{narang2017exploring}. Pruning frequency is set to $1000$. We use the standard scheme where each layer is pruned to the same pruning ratio at every step. 
For \emph{hard pruning} \cite{see2016compression}, pruning is applied to the dense baseline model after training is completed. Retraining is then performed for $10$ epochs. LR and annealing schedule are the same as used for dense baseline.

For inference, beam search is used in order to better approximate $S = \arg \max_{S\,\prime} \: p(S^{\:\prime} \,|\, I)$. Beam size is set to $b = 3$ with no length normalisation. We evaluate the last checkpoint upon completion of training for all the experiments. We denote compression ratio as CR.

\subsection{Dataset}
\label{subsec: Dataset}

The experiments are performed on the popular MS-COCO dataset \cite{lin2014microsoft}. It is a public English captioning dataset which contains $123,287$ images and each image is given at least $5$ captions by different Amazon Mechanical Turk (AMT) workers. As there is no official test split with annotations available, the publicly available split\footnote{\label{footnote: karpathy}http://cs.stanford.edu/people/karpathy/deepimagesent/} in the work of \cite{karpathy2015deep} is used in this work. The split assigns $5,000$ images for validation, another $5,000$ for testing and the rest for training. We reuse the publicly available tokenised captions. Words that occur less than $5$ times are filtered out and sentences longer than $20$ words are truncated.

All the scores are obtained using the publicly available MS-COCO evaluation toolkit\footnote{https://github.com/tylin/coco-caption} , which computes BLEU \cite{papineni2002bleu}, METEOR \cite{banerjee2005meteor}, ROUGE-L \cite{lin2004rouge}, CIDEr \cite{vedantam2015cider} and SPICE \cite{anderson2016spice}. For sake of brevity, we label BLEU-1 to BLEU-4 as B-1 to B-4, and METEOR, ROUGE-L, CIDEr, SPICE as M, R, C, S respectively.


\section{Experiments and Discussion}
\label{sec: Experiments and Discussion}

\subsection{Ablation study}
\label{subsec: Ablation study}

Table \ref{table: Ablation: gating init} shows the effect of various gating initialisation values. From the table, we can see that the best overall performance is achieved when $m$ is set to $5$. Starting the gating parameters at a value of $5$ allows all the captioning parameters $\theta$ to be retained with high probability at the early stages of training, allowing better convergence. This observation is also consistent with the works of \cite{zhu2017prune} and \cite{yu2019playing}, where the authors found that gradual pruning and late resetting can lead to better model performance. Thus, we recommend setting $m = 5.0$.

Table \ref{table: Ablation: sparsity weight} shows the effect of sparsity regularisation weightage $\lambda_{s}$. This is the important hyperparameter that could affect the final sparsity level at convergence. From the results, we can see that low values lead to insufficient sparsity, and higher sparsity target $s_{target}$ requires higher $\lambda_{s}$. For image captioning on MS-COCO, we empirically determined that the heuristic given in Sec. \ref{subsec: Hyperparameters} works sufficiently well for sparsity levels from $80\%$ to $97.5\%$ (see Table \ref{table: LSTM pruning on MS-COCO} and \ref{table: GRU pruning on MS-COCO}.

\begin{table*}[tp]
    \caption{Comparison with dense LSTM baseline and competing methods. \tbf{Bold text} indicates best overall performance. ``\GradualPruning{}'' and ``Hard'' denote methods proposed in \cite{zhu2017prune} and \cite{see2016compression}.}
    \label{table: LSTM pruning on MS-COCO}
    \begin{center}
    \begin{adjustbox}{max width=0.85\linewidth}
    \begin{tabular}{ `l  c ~c ~c  c ~c ~c ~c ~c ~c ~c ~c ~c }
        \toprule
        Approaches              &&\multicolumn{2}{c}{NNZ parameters}&& \multicolumn{8}{c}{MS-COCO test set scores} \\
                                  \cmidrule{3-4}                       \cmidrule{6-13}
        \null                       && Sparsity     & Overall CR    && B-1  & B-2  & B-3  & B-4  & M    & R    & C    & S    \\
        
        \midrule
        Dense LSTM baseline         && 0            & 1 $\times$    && 71.8 & 54.8 & 41.3 & 31.1 & 24.6 & 52.8 & 94.3 & 17.4 \\
        
        \midrule
        \UniformPruning{}           && 0.800        & 5 $\times$    && 71.6 & 54.5 & 41.0 & 30.8 & 24.6 & 52.7 & 93.7 & 17.5 \\
        \DistPruning{}              && \null        & \null         && 71.5 & 54.5 & 40.9 & 30.8 & 24.7 & 52.7 & 93.5 & 17.4 \\
        \BlindPruning{}             && \null        & \null         && 71.5 & 54.7 & 41.2 & 31.1 & 24.7 & 52.7 & 94.2 & 17.5 \\
        \GradualPruning{}           && \null        & \null         && 71.5 & 54.7 & 41.2 & 31.1 & 24.5 & 52.8 & 94.0 & 17.4 \\
   \rbf \Proposed{} ($\lambda_s = 5$) && \null      & \null         && 71.6 & 54.8 & 41.4 & 31.4 & 24.6 & 52.8 & 94.4 & 17.5 \\
        
        \midrule
        \UniformPruning{}           && 0.900        & 10 $\times$   && 70.9 & 53.8 & 40.3 & 30.2 & 24.1 & 52.1 & 90.8 & 16.8 \\
        \DistPruning{}              && \null        & \null         && 70.7 & 53.7 & 40.2 & 30.1 & 24.0 & 52.1 & 90.9 & 16.9 \\
        \BlindPruning{}             && \null        & \null         && 71.1 & 53.9 & 40.4 & 30.3 & 24.2 & 52.2 & 91.8 & 17.2 \\
        \GradualPruning{}           && \null        & \null         && 71.0 & 54.0 & 40.5 & 30.4 & 24.1 & 52.3 & 91.4 & 17.0 \\
   \rbf \Proposed{} ($\lambda_s = 5$) && \null      & \null         && 71.4 & 54.3 & 40.8 & 30.8 & 24.4 & 52.4 & 93.1 & 17.3 \\
        
        \midrule
        \UniformPruning{}           && 0.950        & 20 $\times$   && 69.1 & 51.7 & 38.0 & 27.9 & 22.9 & 50.6 & 83.7 & 15.8 \\
        \DistPruning{}              && \null        & \null         && 68.8 & 51.4 & 37.7 & 27.6 & 22.8 & 50.4 & 83.2 & 15.8 \\
        \BlindPruning{}             && \null        & \null         && 69.5 & 52.5 & 38.9 & 29.0 & 23.3 & 51.3 & 87.0 & 16.3 \\
        \GradualPruning{}           && \null        & \null         && 70.6 & 53.7 & 40.2 & 30.1 & 23.8 & 52.0 & 89.7 & 16.8 \\
   \rbf \Proposed{} ($\lambda_s = 10$) && \null     & \null         && 71.2 & 54.2 & 40.7 & 30.6 & 24.3 & 52.4 & 92.1 & 17.2 \\
        
        \midrule
        \UniformPruning{}           && 0.975        & 40 $\times$   && 66.6 & 48.9 & 35.3 & 25.4 & 21.5 & 48.8 & 75.1 & 14.4 \\
        \DistPruning{}              && \null        & \null         && 65.9 & 48.1 & 34.7 & 25.0 & 21.1 & 48.3 & 72.2 & 14.0 \\
        \BlindPruning{}             && \null        & \null         && 66.9 & 48.9 & 35.3 & 25.6 & 21.6 & 48.9 & 75.9 & 14.6 \\
        \GradualPruning{}           && \null        & \null         && 69.3 & 52.0 & 38.4 & 28.3 & 23.0 & 50.9 & 84.1 & 15.8 \\
   \rbf \Proposed{} ($\lambda_s = 20$) && \null     & \null         && 70.4 & 53.4 & 39.8 & 29.6 & 23.7 & 51.8 & 88.5 & 16.7 \\
       
        \bottomrule
    \end{tabular}
    \end{adjustbox}
    \end{center}
\end{table*}

\begin{table*}[tp]
    \caption{Comparison with dense GRU baseline and competing methods. \tbf{Bold text} indicates best overall performance. ``\GradualPruning{}'' and ``Hard'' denote methods proposed in \cite{zhu2017prune} and \cite{see2016compression}.}
    \label{table: GRU pruning on MS-COCO}
    \begin{center}
    \begin{adjustbox}{max width=0.85\linewidth}
    \begin{tabular}{ `l  c ~c ~c  c ~c ~c ~c ~c ~c ~c ~c ~c }
        \toprule
        Approaches              &&\multicolumn{2}{c}{NNZ parameters}&& \multicolumn{8}{c}{MS-COCO test set scores} \\
                                  \cmidrule{3-4}                       \cmidrule{6-13}
        \null                       && Sparsity     & Overall CR    && B-1  & B-2  & B-3  & B-4  & M    & R    & C    & S    \\
        
        \midrule
        Dense GRU baseline          && 0            & 1 $\times$    && 71.6 & 54.8 & 41.3 & 31.2 & 24.7 & 52.8 & 94.8 & 17.8 \\
        
        \midrule
        \UniformPruning{}           && 0.800        & 5 $\times$    && 70.9 & 53.8 & 40.2 & 30.1 & 24.3 & 52.2 & 92.8 & 17.3 \\
        \DistPruning{}              && \null        & \null         && 71.5 & 54.5 & 40.9 & 30.6 & 24.6 & 52.6 & 93.9 & 17.6 \\
        \BlindPruning{}             && \null        & \null         && 71.3 & 54.2 & 40.7 & 30.6 & 24.6 & 52.6 & 94.1 & 17.7 \\
        \GradualPruning{}           && \null        & \null         && 71.2 & 54.3 & 40.8 & 30.6 & 24.4 & 52.5 & 92.9 & 17.3 \\
   \rbf \Proposed{} ($\lambda_s = 5$) && \null      & \null         && 71.3 & 54.4 & 41.0 & 30.9 & 24.6 & 52.6 & 94.2 & 17.5 \\
        
        \midrule
        \UniformPruning{}           && 0.900        & 10 $\times$   && 70.7 & 53.5 & 39.9 & 29.9 & 23.9 & 51.9 & 90.3 & 16.9 \\
        \DistPruning{}              && \null        & \null         && 70.7 & 53.5 & 40.0 & 30.0 & 24.0 & 51.9 & 90.5 & 17.0 \\
   \rbf \BlindPruning{}             && \null        & \null         && 71.2 & 54.3 & 41.0 & 31.0 & 24.5 & 52.5 & 93.8 & 17.4 \\
        \GradualPruning{}           && \null        & \null         && 70.9 & 53.8 & 40.2 & 30.2 & 24.0 & 52.2 & 91.1 & 16.9 \\
        \Proposed{} ($\lambda_s = 5$) && \null      & \null         && 70.9 & 53.9 & 40.4 & 30.3 & 24.4 & 52.3 & 92.3 & 17.3 \\
        
        \midrule
        \UniformPruning{}           && 0.950        & 20 $\times$   && 68.7 & 51.2 & 37.7 & 27.9 & 22.7 & 50.4 & 83.2 & 15.6 \\
        \DistPruning{}              && \null        & \null         && 68.6 & 51.1 & 37.6 & 27.7 & 22.7 & 50.3 & 83.0 & 15.6 \\
        \BlindPruning{}             && \null        & \null         && 70.6 & 53.4 & 39.8 & 29.7 & 23.8 & 51.8 & 89.3 & 16.7 \\
        \GradualPruning{}           && \null        & \null         && 70.3 & 53.3 & 39.8 & 29.8 & 23.7 & 51.8 & 88.3 & 16.6 \\
   \rbf \Proposed{} ($\lambda_s = 10$) && \null     & \null        && 71.0 & 54.0 & 40.5 & 30.4 & 24.3 & 52.2 & 92.1 & 17.2 \\
        
        \midrule
        \UniformPruning{}           && 0.975        & 40 $\times$   && 66.4 & 48.7 & 35.1 & 25.3 & 21.3 & 48.6 & 75.1 & 14.4 \\
        \DistPruning{}              && \null        & \null         && 66.1 & 48.1 & 34.3 & 24.5 & 21.0 & 48.3 & 72.5 & 14.1 \\
        \BlindPruning{}             && \null        & \null         && 69.1 & 51.5 & 37.8 & 27.7 & 22.6 & 50.5 & 83.2 & 15.6 \\
        \GradualPruning{}           && \null        & \null         && 68.9 & 51.7 & 38.1 & 28.1 & 22.9 & 50.8 & 83.3 & 15.8 \\
   \rbf \Proposed{} ($\lambda_s = 20$) && \null     & \null        && 70.2 & 53.1 & 39.4 & 29.2 & 23.6 & 51.7 & 88.7 & 16.5 \\
       
        \bottomrule
    \end{tabular}
    \end{adjustbox}
    \end{center}
\end{table*}

\subsection{Comparison with RNN pruning methods}
\label{subsec: Comparison pruning methods}

In this section, we provide extensive comparisons of our proposed method with the dense baselines as well as competing methods at multiple sparsity levels. All the models have been verified to have achieved the targeted sparsity levels. From Table \ref{table: LSTM pruning on MS-COCO} and \ref{table: GRU pruning on MS-COCO}, we can clearly see that our proposed end-to-end pruning provides good performance when compared to the dense baselines. This is true even at high pruning ratios of $90\%$ and $95\%$. The relative drops in BLEU-4 and CIDEr scores are only $-1.0\%$ to $-2.9\%$ and $-1.3\%$ to $-2.9\%$ while having $10 - 20 \times$ fewer NNZ parameters. This is in contrast with competing methods whose performance drops are double or even triple compared to ours, especially for LSTM.

The performance advantage provided by end-to-end pruning is even more apparent at the high pruning ratio of $97.5\%$, offering a big $40 \times$ reduction in NNZ parameters. Even though we suffered relative degradations of $-4.8\%$ to $-6.4\%$ in BLEU-4 and CIDEr scores compared to baselines, our performance is still significantly better than the next-closest method which is gradual pruning. On the other hand, the performance achieved by our $80\%$ pruned models are extremely close to that of baselines. Our sparse LSTM model even very slightly outperforms the baseline on some metrics, although we note that the standard deviation for CIDEr score across training runs is around $0.3$ to $0.9$.

Among the competing methods, we can see that gradual pruning usually outperforms hard pruning, especially at high sparsities of $95\%$ and $97.5\%$. That being said, we can see that class-blind hard pruning is able to produce good results at moderate pruning rates of $80\%$ and $90\%$, even outperforming gradual pruning. This is especially true for the GRU captioning model where it outperforms all other methods briefly at $90\%$ sparsity, however we note that its performance on LSTM is generally lower. In contrast, our proposed approach achieves good performance on both LSTM and GRU models.

All in all, these results showcase the strength of our proposed method. Across pruning ratios from $80\%$ to $97.5\%$, our approach consistently maintain relatively good performance when compared to the dense baselines while outperforming magnitude-based gradual and hard pruning methods in most cases.

\begin{table*}[tp]
    \caption{Comparison with dense LSTM and GRU baselines after CNN fine-tuning.}
    \label{table: CNN fine-tune on MS-COCO}
    \begin{center}
    \begin{adjustbox}{max width=0.85\linewidth}
    \begin{tabular}{ `l  c ~c ~c  c ~c ~c ~c ~c ~c ~c ~c ~c }
        \toprule
        Approaches          &&\multicolumn{2}{c}{NNZ parameters}&& \multicolumn{8}{c}{MS-COCO test set scores} \\
                              \cmidrule{3-4}                       \cmidrule{6-13}
        \null                   && RNN size     & Overall CR    && B-1  & B-2  & B-3  & B-4  & M    & R    & C     & S    \\
        
        \midrule
        H-LSTM \cite{dai2018grow} + GP \cite{dai2019nest} && 394 K &\null &&\null &\null &\null &\null &\null &\null &  95.4 &\null \\
        \null                   && 163 K        & \null         &&\null &\null &\null &\null &\null &\null &  93.3 &\null \\
        
        \midrule
        Dense LSTM baseline     && 2.62 M       &  1 $\times$   && 73.8 & 57.5 & 44.0 & 33.5 & 25.8 & 54.7 & 102.6 & 18.7 \\
        
        \Proposed{}             &&  725 K       &  5 $\times$   && 73.3 & 56.8 & 43.1 & 32.7 & 25.6 & 54.1 & 100.9 & 18.6 \\
        \null                   &&  402 K       & 10 $\times$   && 73.5 & 57.1 & 43.5 & 33.2 & 25.5 & 54.2 & 101.3 & 18.5 \\
        \null                   &&  205 K       & 20 $\times$   && 73.6 & 57.2 & 43.6 & 33.2 & 25.5 & 54.2 & 100.8 & 18.4 \\
        \null                   &&  101 K       & 40 $\times$   && 73.4 & 57.0 & 43.3 & 32.8 & 25.2 & 54.0 & 100.0 & 18.2 \\
        (Beam size = 2)         &&  101 K       & 40 $\times$   && 73.5 & 56.9 & 43.0 & 32.2 & 25.2 & 53.8 &  98.9 & 18.2 \\
        
        \midrule
        Dense GRU baseline      && 1.97 M       &  1 $\times$   && 73.4 & 56.6 & 42.9 & 32.4 & 25.6 & 54.0 & 100.6 & 18.5 \\
        
        \Proposed{}             &&  589 K       &  5 $\times$   && 73.0 & 56.3 & 42.7 & 32.3 & 25.5 & 54.0 &  99.8 & 18.5 \\
        \null                   &&  361 K       & 10 $\times$   && 73.2 & 56.7 & 43.0 & 32.6 & 25.5 & 54.1 & 100.0 & 18.5 \\
        \null                   &&  185 K       & 20 $\times$   && 73.3 & 56.8 & 43.2 & 32.8 & 25.5 & 54.2 & 100.9 & 18.3 \\
        \null                   &&   89 K       & 40 $\times$   && 73.1 & 56.6 & 42.8 & 32.3 & 25.1 & 53.8 &  98.7 & 18.2 \\
       
        \bottomrule
    \end{tabular}
    \end{adjustbox}
    \end{center}
\end{table*}

\begin{table*}[tp]
    \caption{Comparison of large-sparse and small-dense LSTM models.}
    \label{table: Sparse vs dense}
    \begin{center}
    \begin{adjustbox}{max width=0.85\linewidth}
    \begin{tabular}{ `l  c ~c ~c ~c  c ~c ~c ~c ~c ~c ~c ~c ~c }
        \toprule
        Models          &&\multicolumn{3}{c}{NNZ parameters}    && \multicolumn{8}{c}{MS-COCO test set scores} \\
                          \cmidrule{3-5}                           \cmidrule{7-14}
        \null           && Sparsity & Params.   & Overall CR    && B-1  & B-2  & B-3  & B-4  & M    & R    & C    & S    \\
        
        \midrule
        LSTM-M          && 0        & 11.9 M    & 1 $\times$    && 71.8 & 54.8 & 41.3 & 31.1 & 24.6 & 52.8 & 94.3 & 17.4 \\ 
        
        \midrule
        LSTM-S          && 0        & 2.4 M     & 5 $\times$    && 69.6 & 52.5 & 38.5 & 28.1 & 22.7 & 50.9 & 82.7 & 15.7 \\ 
        
        \midrule
        LSTM-M          && 0.800    & 2.4 M     & 5 $\times$    && 71.6 & 54.8 & 41.4 & 31.4 & 24.6 & 52.8 & 94.4 & 17.5 \\ 
        \null           && 0.900    & 1.2 M     & 10 $\times$   && 71.4 & 54.3 & 40.8 & 30.8 & 24.4 & 52.4 & 93.1 & 17.3 \\ 
        \null           && 0.950    & 0.6 M     & 20 $\times$   && 71.2 & 54.2 & 40.7 & 30.6 & 24.3 & 52.4 & 92.1 & 17.2 \\ 
        \null           && 0.975    & 0.3 M     & 40 $\times$   && 70.4 & 53.4 & 39.8 & 29.6 & 23.7 & 51.8 & 88.5 & 16.7 \\ 
       
        \bottomrule
    \end{tabular}
    \end{adjustbox}
    \end{center}
\end{table*}

\begin{table*}[tp]
    \caption{Comparison on caption uniqueness and length with dense baselines on MS-COCO test set.}
    \label{table: Uniqueness on MS-COCO}
    \begin{center}
    \begin{adjustbox}{max width=0.85\linewidth}
    \begin{tabular}{ `l  c ~c ~c  c ~c ~c  c ~c ~c }
        \toprule
        Approaches              &&\multicolumn{2}{c}{NNZ parameters}&& \multicolumn{2}{c}{Before fine-tune} && \multicolumn{2}{c}{After fine-tune} \\
                                  \cmidrule{3-4}                       \cmidrule{6-7}                          \cmidrule{9-10}
        \null                       && Sparsity     & Overall CR    && Unique (\%)  & Average length && Unique (\%)  & Average length \\
        
        \midrule
        Dense LSTM baseline         && 0            &  1 $\times$   && 42.1 & 9.09 && 46.0 & 9.07 \\
        
        \Proposed{}                 && 0.800        &  5 $\times$   && 42.5 & 9.11 && 45.3 & 9.12 \\
        \null                       && 0.900        & 10 $\times$   && 41.9 & 9.07 && 46.8 & 9.09 \\
        \null                       && 0.950        & 20 $\times$   && 41.9 & 9.05 && 47.0 & 9.05 \\
        \null                       && 0.975        & 40 $\times$   && 44.4 & 8.99 && 48.4 & 8.97 \\
        
        \midrule
        Dense GRU baseline          && 0            &  1 $\times$   && 42.4 & 9.15 && 46.9 & 9.14 \\
        
        \Proposed{}                 && 0.800        &  5 $\times$   && 43.1 & 9.13 && 47.3 & 9.16 \\
        \null                       && 0.900        & 10 $\times$   && 43.0 & 9.09 && 46.2 & 9.13 \\
        \null                       && 0.950        & 20 $\times$   && 42.7 & 9.07 && 46.9 & 9.06 \\
        \null                       && 0.975        & 40 $\times$   && 42.0 & 8.94 && 49.1 & 8.98 \\
        
        \bottomrule
    \end{tabular}
    \end{adjustbox}
    \end{center}
\end{table*}

\begin{figure*}[tp]
    \begin{center}
    \begin{subfigure}{.95\linewidth}
        \centering
        \includegraphics[keepaspectratio=true, scale=0.125]{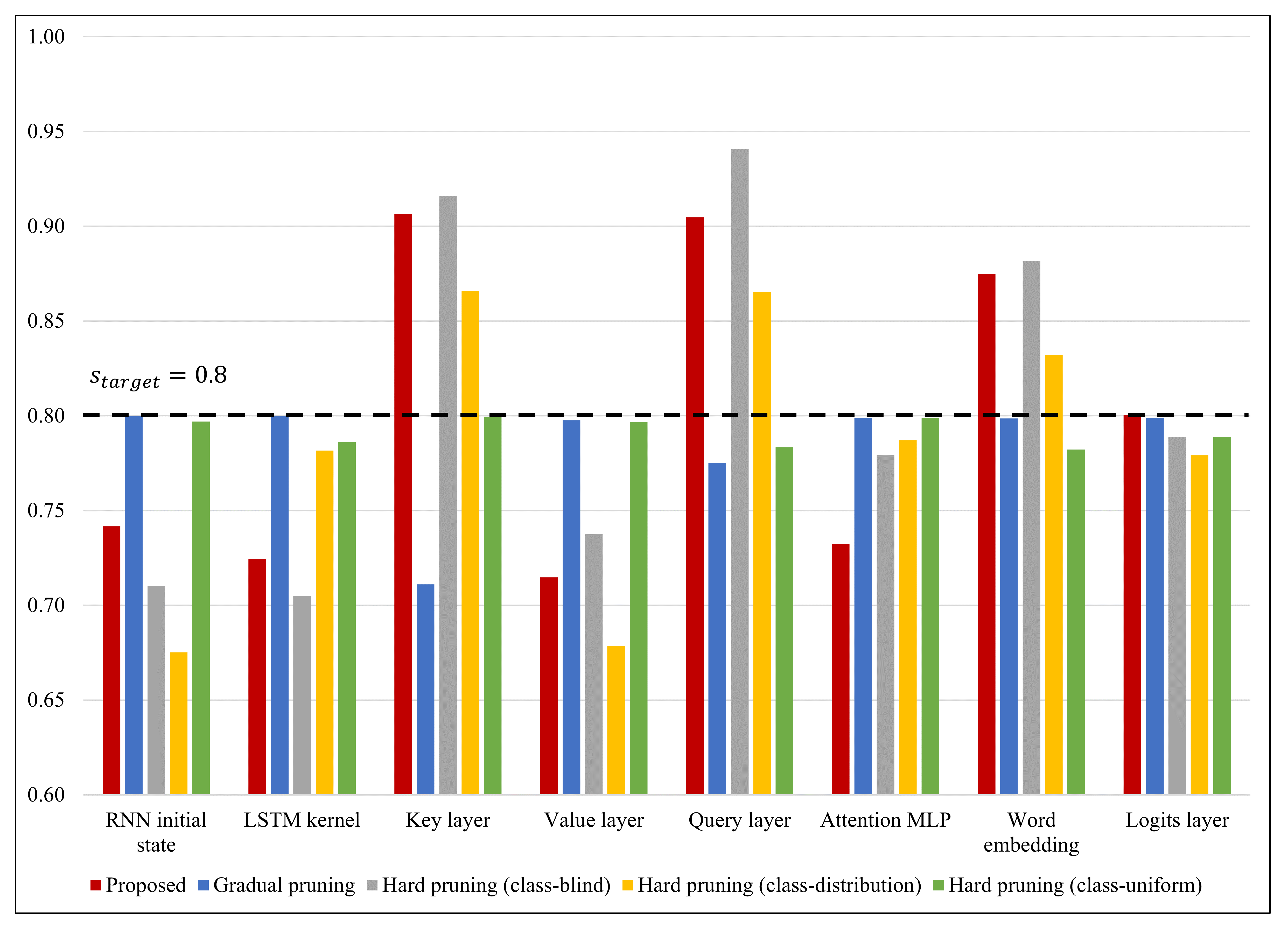}
        \caption{Pruning ratios at $s_{target} = 0.8$}
        \label{fig: Layerwise compare 0.8}
    \end{subfigure}
    \hfill
    \begin{subfigure}{.95\linewidth}
        \centering
        \includegraphics[keepaspectratio=true, scale=0.125]{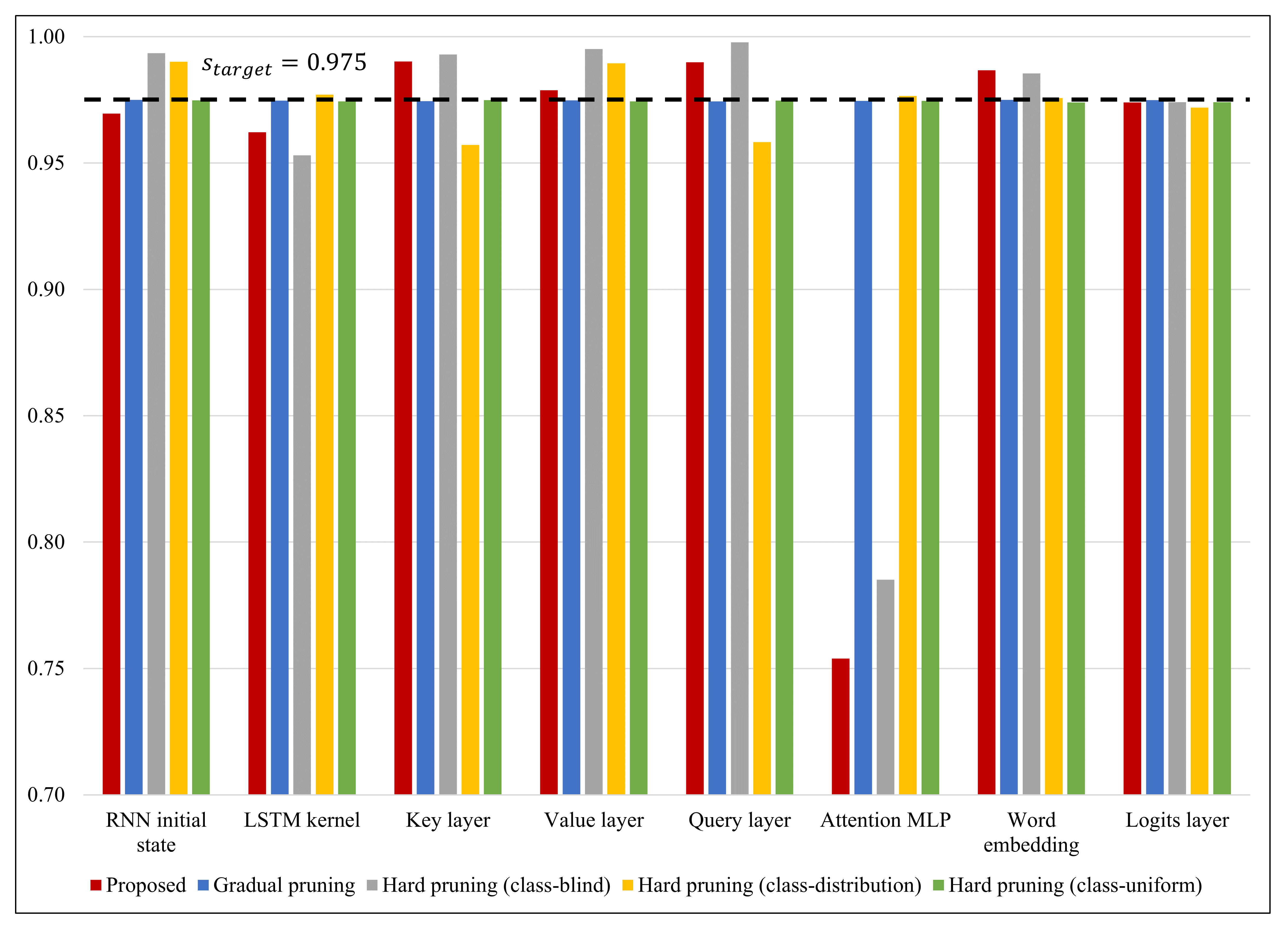}
        \caption{Pruning ratios at $s_{target} = 0.975$}
        \label{fig: Layerwise compare 0.975}
    \end{subfigure}
    \end{center}
    \caption{Layer-wise comparison of final sparsity levels. Best viewed in colour.}
    \label{fig: Layerwise compare}
\end{figure*}

\subsection{Effect of fine-tuning}
\label{subsec: Effect of fine-tuning}

In this section, we investigate the potential impact of fine-tuning the entire captioning model in an end-to-end manner. From Table \ref{table: CNN fine-tune on MS-COCO}, we can see that model fine-tuning has a performance-recovering effect on the sparse models. This phenomenon is especially apparent on very sparse models with sparsity at $97.5\%$. On both LSTM and GRU models, the drops in performance suffered due to pruning have mostly reduced except for LSTM at $80\%$ sparsity.

Notably, all the pruned models have remarkably similar performance from $80\%$ sparsity up until $97.5\%$. The score gap between dense and sparse GRU models are exceedingly small, ranging from $+1.2\%$ to $-1.9\%$ for both BLEU-4 and CIDEr. For LSTM models, even though the score gap is slightly larger at $-0.9\%$ to $-2.5\%$ on both BLEU-4 and CIDEr, it is still considerably smaller than without CNN fine-tuning (Table \ref{table: LSTM pruning on MS-COCO}).

These results suggest that the Inception-V1 CNN pre-trained on ImageNet is not optimised to provide useful features for sparse decoders. As such, end-to-end fine-tuning together with sparse decoder allows features extracted by the CNN to be adapted where useful semantic information can be propagated through surviving connections in the decoder.

We also provided compression and performance comparison with the closely related work of \cite{dai2018grow} who utilised GP \cite{dai2019nest} method to produce sparse H-LSTM for image captioning. For fairness, we also provide scores obtained at CR of $40 \times$ using beam size of $2$ instead of $3$. From the table, we can see that at overall CR of $20 \times$ to $40 \times$, our sparse models are able to outperform H-LSTM with lower NNZ parameters. This indicates that the effectiveness of our one-shot approach is at least comparable to the iterative process of grow-and-prune.

\subsection{Large-sparse versus small-dense}
\label{subsec: Sparse vs dense}

In this section, we show that a large sparse LSTM image captioning model produced via end-to-end pruning is able to outperform a smaller dense LSTM trained normally. The small-dense model denoted as \emph{LSTM-S} has a word embedding size of $q = 64$ dimensions, LSTM size of $r = 128$ units and finally attention MLP size of $a = 96$ units.
The results are given in Table \ref{table: Sparse vs dense}. From the results, we can see that the small-dense model with $5 \times$ fewer parameters performs considerably worse than all the large-sparse models \emph{LSTM-M} across the board.

Notably, we can see that the large-sparse \emph{LSTM-M} model with $40 \times$ fewer NNZ parameters still managed to outperform \emph{LSTM-S} with a considerable margin. At equal NNZ parameters, the large-sparse model comfortably outperforms the small-dense model. This showcases further the strength of model pruning and solidifies the observations made in works on RNN pruning \cite{narang2017exploring,zhu2017prune}.

\subsection{Caption uniqueness and length}
\label{subsec: Caption uniqueness}

In this section, we explore the potential effects of our proposed end-to-end pruning on the uniqueness and length of the generated captions. As pruning reduces the complexity and capacity of the decoder considerably, we wish to see if the sparse models show any signs of training data memorisation and hence potentially overfitting. In such cases, uniqueness of the generated captions would decrease as the decoder learns to simply repeat captions available in the training set. A generated caption is considered to be unique if it is not found in the training set.

From Table \ref{table: Uniqueness on MS-COCO}, we can see that despite the heavy reductions in NNZ parameters, the uniqueness of generated captions have not decreased. On the contrary, more unseen captions are being generated at higher levels of sparsity and compression. On the other hand, we can see that the average lengths of generated captions peaked at $80\%$ sparsity in most cases and then decrease slightly as sparsity increase. That being said, the reductions in caption length are minimal ($+0.5\%$ to $-2.3\%$) considering the substantial decoder compression rates of up to $40 \times$.

Together with the good performance shown in Table \ref{table: LSTM pruning on MS-COCO} and \ref{table: GRU pruning on MS-COCO}, these results indicate that our approach is able to maintain both the variability of generated captions and their quality as measured by the metric scores.

\subsection{Layer-wise sparsity comparison}
\label{subsec: Layer-wise sparsity}

Finally, we visualise the pruning ratio of each decoder layers when pruned using the different methods listed in Sec. \ref{subsec: Comparison pruning methods}. Among the approaches, both gradual and class-uniform pruning produces the same sparsity level across all the layers. To better showcase the differences in layer-wise pruning ratios, we decided to visualise two opposite ends in which the first has a relatively moderate sparsity of $80\%$ while the other has a high sparsity of $97.5\%$.

In both Fig. \ref{fig: Layerwise compare 0.8} and \ref{fig: Layerwise compare 0.975}, we denote the decoder layers as follows: ``RNN initial state'' refers to $W_I$ in Eq. \ref{eq: LSTM init}; ``LSTM kernel'' is the concatenation of all gate kernels in LSTM (\ie{} input, output, forget, cell); ``Key'', ``Value'' and ``Query'' layers refer to projection layers in the attention module (see \cite{vaswani2017attention} for details); ``Attention MLP'' is the second layer of the 2-layer attention MLP; and finally ``Word'' and ``Logits'' refer to the word embedding matrix $E_w$ in Eq. \ref{eq: RNN input} and $E_o$ in Eq. \ref{eq: RNN output} respectively.

From the figures, we can see that our proposed pruning method consistently prune ``attention MLP'' layer the least. This is followed by ``LSTM kernel'' and ``Value'' layers where they generally receive lesser pruning compared to others. On the flip side, ``Key'' and ``Query'' layers were pruned most heavily at levels often exceeding the targeted pruning rates. Finally, ``Word embedding'' consistently receives more pruning than ``Logits layer''. This may indicate that there exists substantial information redundancy in the word embeddings matrix as noted in works such as \cite{shi2018structured,tan2019comic,shu2017compressing}.


\section{Conclusion and Future Work}
\label{sec: Conclusion and Future Work}

In this work, we have investigated the effectiveness of model weight pruning on the task of image captioning with visual attention. In particular, we proposed an end-to-end pruning method that performs considerably better than competing methods at maintaining captioning performance while maximising compression rate. Our single-shot approach is simple and fast to use, provides good performance, and its sparsity level is easy to tune.
Moreover, we have demonstrated by pruning decoder weights during training, we can find sparse models that performs better than dense counterparts while significantly reducing model size.

Our results pave the way towards deployment on mobile and embedded devices due to their small size and reduced memory requirements. In the future, we wish to investigate the generalisation capability of end-to-end pruning when applied on Transformer models \cite{vaswani2017attention}. We would also like to extend our method to other CV and NLP tasks including image classification, language modelling and natural language translation.


\singlespacing

\renewcommand\refname{Bibliography}
\bibliography{ref_captioning,ref_nlp,ref_compact,ref_misc}

\end{document}